\documentclass[a4,11pt]{article}
\usepackage[T1]{fontenc}
\usepackage{graphicx}
\usepackage{amsmath}
\usepackage{booktabs}
\usepackage{float}
\usepackage{changepage}
\usepackage{geometry}
\usepackage{url}
\begin{document}
\setlength{\intextsep}{6pt}
\setlength{\textfloatsep}{6pt}
\setlength{\floatsep}{6pt}
\setlength{\abovecaptionskip}{3pt}
\setlength{\belowcaptionskip}{3pt}
\title{Enhancing Automated Machine Learning via \emph{Homogeneous} Train–-Test Splitting Methods}
%
%
\author{Yearn Tan Yin Tze and
Charles Grellois\\School of Computer Science, University of Sheffield, U.K.}
%
\date{\emph{Accepted at UKCI'26 -- pre-review version, May 2026}}

\maketitle              
\begin{abstract}
Accurate model evaluation in machine learning depends critically on how datasets are split into training and testing subsets. Standard random splitting assumes that both partitions share the same underlying distribution, an assumption often violated in datasets with class imbalance, natural clustering, or spatial autocorrelation. This paper investigates the role of statistical similarity in train--test splitting and its consequences for AutoML model evaluation. Five established strategies are compared across fifteen UCI benchmark datasets: random splitting, stratified sampling, Kennard--Stone, Duplex, and SPXY. Similarity is assessed using chi-square, Kolmogorov--Smirnov, and Maximum Mean Discrepancy (MMD) tests. Geometry-based methods consistently produce near-zero MMD scores, introducing instability into downstream performance estimates. The proposed Optimised-Distribution method treats similarity as an explicit optimisation objective and achieves the highest mean MMD similarity, 89.0\%, across all strategies evaluated.
\end{abstract}
\section{Introduction}

Machine learning (ML) has become integral to decision-making across numerous industries, including healthcare, security, and chemical process engineering~\cite{sarker2021machine}. The reliability of ML systems depends heavily on accurate performance estimation, which is typically conducted by splitting a dataset into training and testing subsets~\cite{hastie2009elements}. A foundational assumption behind random train--test splitting is that both subsets are statistically representative of the same underlying data distribution. When this assumption is violated because of dataset shift, class imbalance, or spatial autocorrelation~\cite{moreno2012unifying,salazar2022fair}, performance estimates become unreliable~\cite{moreno2012unifying}.

In Automated Machine Learning (AutoML), this problem is particularly critical. AutoML systems automatically select algorithms, preprocess data, and optimise hyperparameters. If data are partitioned in a biased or inconsistent manner, the AutoML pipeline may overfit to training peculiarities and yield deceptively strong evaluation metrics, undermining model selection and generalisation~\cite{hutter2019automated}.

Traditional approaches such as stratified sampling preserve class proportions but do not guarantee feature-space similarity. Geometry-based methods like the Kennard--Stone (KS) algorithm~\cite{kennard1969computer} and SPXY~\cite{galvao2005method} maximise coverage of the feature space but, by design, create training sets that are distributionally dissimilar from test sets. No prior work has systematically evaluated these trade-offs using a unified distributional similarity framework across diverse tabular benchmark datasets within an AutoML context.

This paper addresses that gap. Our main contributions are:
\begin{enumerate}
    \item A systematic comparative evaluation of five established splitting strategies using chi-square, KS, and MMD similarity tests across fifteen UCI datasets
    \item Evidence that geometry-based methods consistently underperform random and stratified splitting on multivariate distributional similarity
    \item Refined Optimised-Distribution splitting method that explicitly maximises statistical similarity through iterative similarity-guided sample swapping.
\end{enumerate}

\section{Background and Related Work}

\subsection{The Train--Test Split Problem}

Standard random splitting assigns samples to training and test sets with uniform probability. While computationally efficient, it does not guarantee that the resulting partitions are statistically similar~\cite{kohavi1995study}. This phenomenon, termed distributional shift, can arise from class imbalance, natural clustering, or structured spatial patterns in the data. Consequences include over-optimistic or misleadingly pessimistic performance estimates, and reduced model reliability at deployment~\cite{moreno2012unifying}.

\subsection{Existing Splitting Strategies}

Stratified sampling preserves target class proportions across partitions, making it particularly relevant for imbalanced classification tasks~\cite{xu2018splitting}. However, it does not ensure feature-space coverage beyond the stratification variable. The Kennard--Stone (KS) algorithm~\cite{kennard1969computer} is a deterministic method that iteratively selects training samples by maximising feature-space coverage via farthest-point sampling. The test set consists of the remaining samples, which are not directly optimised.

The SPXY algorithm~\cite{galvao2005method} extends KS by incorporating both feature space $X$ and response space $Y$ in the distance criterion, making it useful when target range diversity is important. The Duplex algorithm~\cite{snee1977validation} alternately assigns the most spatially distant unselected sample pairs to training and test sets, aiming to make both partitions representative of the full feature space.

Despite these advances, a key limitation persists: none of these methods treats distributional similarity between train and test sets as an explicit optimisation objective. KS and SPXY focus on training-set coverage; Duplex aims for coverage in both sets but does not directly minimise a distributional divergence measure.

\subsection{Statistical Similarity Measures}

Three complementary tests are used in this work. The Chi-Square Test of Homogeneity assesses whether two groups differ in their categorical variable distributions~\cite{franke2012chi}. Cramér's $V$~\cite{cramer1946mathematical} is derived from the chi-square statistic to quantify effect size. The Kolmogorov--Smirnov (KS) test is a non-parametric test comparing continuous empirical distribution functions~\cite{massey1951kolmogorov}. Maximum Mean Discrepancy (MMD)~\cite{gretton2012kernel} is a kernel-based two-sample test that measures distributional distance in a reproducing kernel Hilbert space (RKHS), enabling multivariate comparison of mixed-type data via a combined RBF/delta kernel. PSI and log-linear analysis were considered but excluded due to their variable-independence assumptions and exponential computational complexity, respectively.

\subsection{AutoML Frameworks}

PyCaret was selected as the primary AutoML framework because it provides a practical balance between automation, computational efficiency, and compatibility with custom train--test splitting strategies~\cite{dissanayake2025survey}. Its pipeline supports automated preprocessing, model comparison, hyperparameter tuning, and evaluation, while still allowing externally generated splits to be incorporated into the experimental workflow. Scikit-learn~\cite{pedregosa2011scikit} was used alongside PyCaret~\cite{ali2020pycaret,dissanayake2025survey} for implementing splitting procedures, preprocessing utilities, and kernel-based similarity computations. Table~\ref{tab:automl_comparison} summarises the AutoML tools considered during framework selection.
\begin{table}[H]
\centering
\begin{tabular}{|l|l|l|}
\hline
\textbf{Tool} & \textbf{Runtime} & \textbf{Suitability} \\
\hline
PyCaret & Moderate & Suitable balance of automation/efficiency/flexibility \\
\hline
AutoGluon & Fast & Strong predictive performance, but GPU\\&& requirements limited repeatability \\
\hline
Auto-sklearn & Moderate--Long & Flexible, but more computationally demanding\\&& than PyCaret \\
\hline
H2O AutoML & Long & Longer training times and occasional failed runs \\
\hline
TPOT & Longest & Low completion rate in preliminary testing; \\&&impractical for repeated experiments \\
\hline
\end{tabular}
\caption{Comparison of evaluated AutoML tools.}
\label{tab:automl_comparison}
\end{table}
\vspace{-30pt}

\section{Methodology}

\subsection{Research Questions}

This work is guided by five research questions:
\begin{itemize}
    \item \textbf{RQ1:} To what extent do common train--test split methods differ in distributional similarity?
    \item \textbf{RQ2:} Do advanced methods produce more representative splits than random splitting?
    \item \textbf{RQ3:} Can a refined method be developed that explicitly optimises for statistical similarity?
    \item \textbf{RQ4:} Does improved similarity lead to more reliable AutoML performance estimates?
    \item \textbf{RQ5:} How do dataset properties moderate the similarity--performance relationship?
\end{itemize}

\subsection{Datasets}

Fifteen classification datasets were sourced from the UCI Machine Learning Repository, selected to represent diversity in size, dimensionality, and feature type: numerical, categorical, and mixed~\cite{dua2017uci}. Dataset sizes range from 150 instances, Iris, UCI 53, to 253,680 instances, CDC Diabetes Health Indicators, UCI 891, and feature dimensionality from 4 to 169. All experiments use a fixed 70:30 train--test split ratio. A summary is presented in Table~\ref{tab:datasets}.

\begin{table}[htbp]
\centering
\caption{Benchmark datasets from the UCI Machine Learning Repository.}
\label{tab:datasets}
\begin{tabular}{|r|r|r|r|l|}
\toprule
\textbf{UCI ID} & \textbf{Instances} & \textbf{Features} & \textbf{Type} & \textbf{Dataset Name} \\
\midrule
17  & 569     & 31  & Numerical   & Breast Cancer Wisconsin \\
42  & 214     & 9   & Numerical   & Glass Identification \\
45  & 303     & 13  & Mixed       & Heart Disease \\
53  & 150     & 4   & Numerical   & Iris \\
74  & 476     & 169 & Mixed       & Musk Version 1 \\
75  & 6,598   & 166 & Numerical   & Musk Version 2 \\
159 & 19,020  & 10  & Numerical   & Magic Gamma Telescope \\
222 & 45,211  & 16  & Mixed       & Bank Marketing \\
264 & 14,980  & 14  & Numerical   & EEG Eye State \\
367 & 102,944 & 115 & Categorical & Dota2 Games Results \\
419 & 292     & 20  & Categorical & Autistic Spectrum Disorder \\
519 & 299     & 12  & Mixed       & Heart Failure Clinical Records \\
529 & 520     & 16  & Mixed       & Early Stage Diabetes Risk \\
857 & 200     & 28  & Categorical & Chronic Kidney Disease \\
891 & 253,680 & 21  & Mixed       & CDC Diabetes Health Indicators \\
\bottomrule
\end{tabular}
\end{table}
\vspace{-15pt}

\subsection{Splitting Strategies}

Six strategies are evaluated. Random splitting assigns samples uniformly at random as the baseline. Stratified sampling preserves target class proportions using scikit-learn's \texttt{train\_test\_split} with stratification. The Kennard--Stone algorithm iteratively selects training samples that maximise coverage of the feature space. Numerical features are standardised and categorical features are one-hot encoded prior to computing pairwise distances. A fast variant caps the working set at 500 samples for large datasets. SPXY applies stratified farthest-point selection independently within each class stratum. Duplex alternately assigns the most distant sample pairs to training and test sets within each class. Optimised-Distribution, the proposed method, is described in Section~\ref{sec:optimised_distribution}.

\subsection{Evaluation Framework}

Statistical similarity is evaluated using three tests, each converted to a unified 0--100\% scale where higher values indicate greater similarity. For datasets with 1--3 categorical features, the chi-square test with Cramér's $V$ is used:
\begin{equation}
    \text{Similarity}_{\chi^2} = (1 - V) \times 100.
\end{equation}

For datasets with 1--3 continuous features, the KS test is used:
\begin{equation}
    \text{Similarity}_{\text{KS}} = (1 - D) \times 100,
\end{equation}
where $D$ is the KS statistic.

For high-dimensional or mixed-type datasets, MMD with a mixed RBF/delta kernel is applied using a permutation test with 50 resamples and z-score normalisation:
\begin{equation}
    \text{Similarity}_{\text{MMD}} = \exp(-\max(z, 0)) \times 100.
\end{equation}

A threshold of 90\% is treated as the target for a high-quality split.

Downstream performance is evaluated using PyCaret on the resulting partitions. For each dataset--method combination, a model comparison is run over logistic regression, decision tree, random forest, gradient boosting, extra trees, k-nearest neighbours, naïve Bayes, and SVM. The best-performing model by weighted F1 is tuned and evaluated on the held-out test set. Both accuracy and weighted F1 are reported; weighted F1 is the primary metric given class imbalance in several datasets.

\subsection{Proposed Optimised-Distribution Splitting}
\label{sec:optimised_distribution}

The proposed method begins with a stratified split, then iteratively refines it by swapping candidate samples between training and test partitions within each class. A swap is accepted only when it reduces the composite objective:
\begin{equation}
    \mathcal{L}
    =
    \text{MMD}
    +
    \lambda \sum_c
    \left|
    \hat{p}_{\text{train}}(c)
    -
    \hat{p}_{\text{test}}(c)
    \right|,
\end{equation}
where MMD is averaged over $n_{\text{repeats}} = 5$ independent subsamples of up to 500 samples per partition, $\lambda = 2.0$ is the class-proportion penalty weight, and $\hat{p}(c)$ is the empirical class proportion. Optimisation runs for at most $\texttt{max\_iter} = 50$ iterations with $\texttt{candidates\_per\_iter} = 20$ swaps evaluated per iteration. The method terminates early when no improving swap is found.

\section{Results and Discussion}
\label{sec:results}
\subsection{Statistical Similarity}

Table~\ref{tab:stat_similarity_scores} presents mean similarity scores averaged across all applicable test--dataset combinations for each method. The Optimised-Distribution method achieves the highest mean MMD similarity, followed by Stratified and Random. Kennard--Stone and SPXY score near zero on MMD across nearly all datasets.

\begin{table}[H]
\centering
\begin{tabular}{|l|c|c|c|c|}
\hline
\textbf{Method} & \textbf{Avg.} & \textbf{Avg.} & \textbf{Avg. Uni-} & \textbf{Avg.} \\
 & \textbf{$\chi^2$ (\%)} & \textbf{KS (\%)} & \textbf{variate (\%)} & \textbf{MMD (\%)} \\
\hline
Random & 88.0 & 93.7 & 91.3 & 78.9 \\
Stratified & 87.7 & 94.1 & 91.4 & 80.2 \\
Kennard--Stone & 78.9 & 79.7 & 79.4 & 6.9 \\
SPXY & 83.3 & 85.9 & 84.8 & 9.7 \\
Duplex & 86.9 & 89.4 & 88.4 & 30.0 \\
Optimised-Distribution & 88.2 & 94.7 & 92.0 & 89.0 \\
\hline
\end{tabular}

\caption{Average statistical similarity scores across all methods.}
\label{tab:stat_similarity_scores}
\end{table}
\vspace{-10pt}

The near-zero MMD scores for Kennard--Stone and SPXY reflect their space-filling design: both methods deliberately bias training sets towards extreme and diverse samples, producing training distributions that are fundamentally different from the test distribution. This is an intentional asymmetry that scores poorly under MMD. A notable exception is UCI 529, where Kennard--Stone achieves 100\% MMD similarity, likely because the dataset's moderate size, 520 instances and 16 features, limits how much distributional divergence the space-filling selection can introduce.

Duplex produces intermediate results: its MMD scores range widely across datasets, from 0.0\% on UCI 159 to 100\% on UCI 53, 74, and 857, reflecting sensitivity to dataset structure. Among high-similarity methods, random and stratified splitting meet the 90\% KS threshold in the majority of cases. Notably, on UCI 891, UCI 264, and UCI 222, stratified splitting achieves low MMD scores, 28.7\%, 18.6\%, and 20.4\% respectively, despite near-perfect KS and chi-square scores, demonstrating that class-stratification can introduce multivariate feature-space imbalance when class boundaries do not align with the underlying feature distribution.

The Optimised-Distribution method achieves 100\% MMD similarity on 13 of the 15 datasets, with the largest improvements over stratified splitting observed on UCI 264, from 18.6\% to 100\%, and UCI 367, from 47.8\% to 100\%. On UCI 891 and UCI 222, improvements are smaller, to 11.9\% and 23.5\%, constrained by the optimisation budget on these very large datasets.

\subsection{Downstream Model Performance}

Table~\ref{tab:downstream_performance} reports average accuracy and weighted F1 across all datasets for each splitting method. The Optimised-Distribution method achieves the highest mean F1, followed by SPXY, Stratified, Kennard--Stone, Random, and Duplex. These aggregate differences are modest, in part because several datasets, including UCI 419, UCI 529, and UCI 857, yield near-perfect scores under all methods, creating a ceiling effect.

\begin{table}[H]
\centering
\begin{tabular}{|l|r|r|}
\hline
\textbf{Method} & \textbf{Mean Accuracy} & \textbf{Mean Weighted F1} \\
\hline
Random              & 0.864 & 0.860 \\
Stratified          & 0.870 & 0.866 \\
Kennard--Stone      & 0.873 & 0.865 \\
SPXY                & 0.875 & 0.868 \\
Duplex              & 0.850 & 0.849 \\
Optimised           & 0.877 & 0.874 \\
\hline
\end{tabular}
\caption{Mean downstream AutoML performance across 15 datasets.}
\label{tab:downstream_performance}
\end{table}
\vspace{-10pt}

The most consequential differences arise on harder, non-trivially separable datasets. UCI 264 shows the largest single gap: Kennard--Stone yields an F1 of 0.701 against 0.964 for both Random and Stratified, consistent with its near-zero MMD similarity on that dataset. On UCI 159, SPXY's F1 drops to 0.756 versus 0.879 for Random, and on UCI 891, SPXY yields 0.706 F1 compared to 0.796 for Random. These patterns reflect SPXY's tendency to concentrate extreme samples in training, leaving an unrepresentative test set.

However, the relationship is not uniformly negative for geometry-based methods. On UCI 519, SPXY and Kennard--Stone achieve F1 scores of 0.909 and 0.887 versus 0.760 for Random, and on UCI 891 Kennard--Stone achieves the highest accuracy of all methods, 0.927. This suggests that for certain datasets, broader training-set coverage can be beneficial for model learning even when it reduces test representativeness.

\subsection{Similarity--Performance Relationship}

\begin{figure}[H]
\centering
\includegraphics[width=0.48\textwidth]{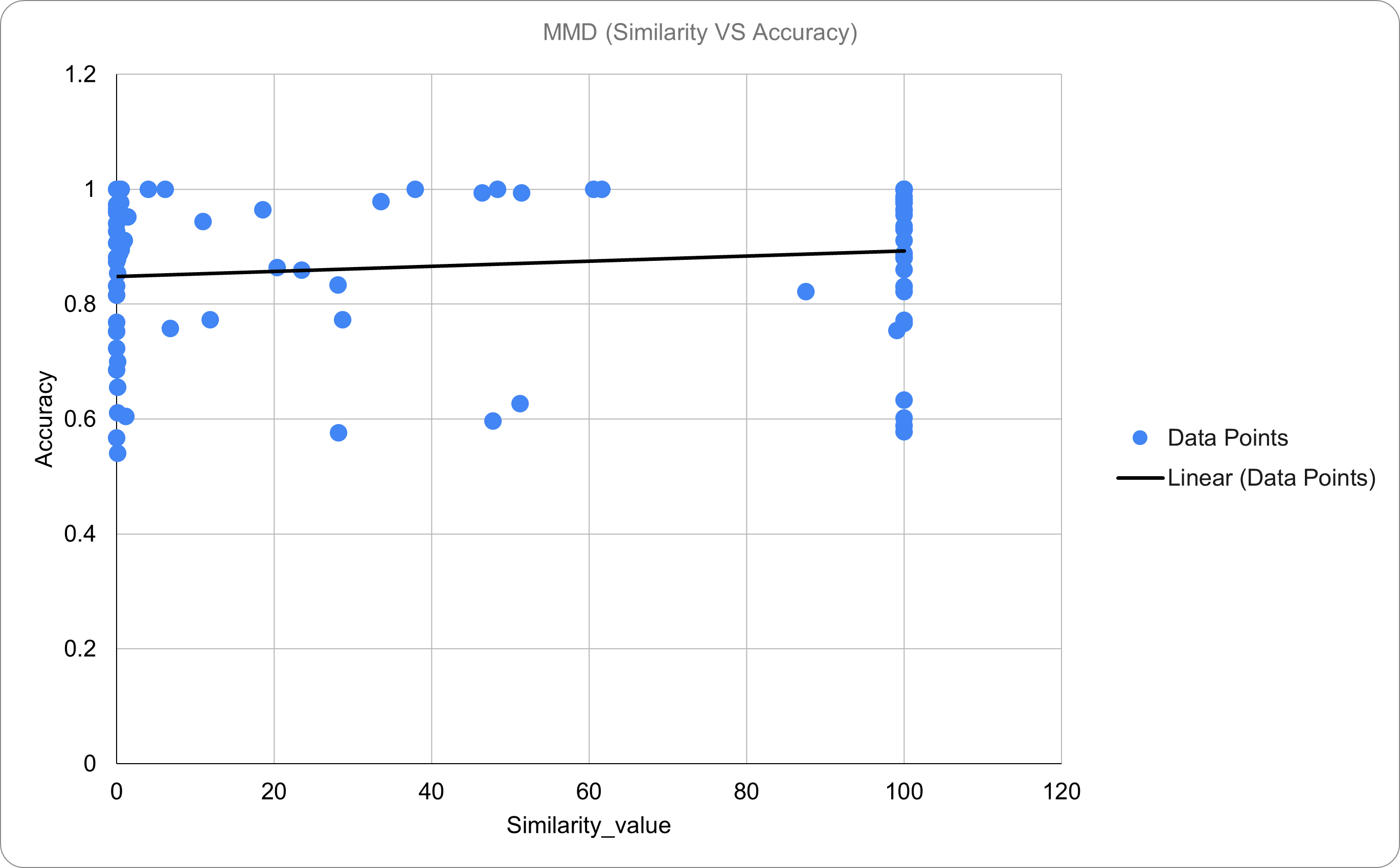}
\includegraphics[width=0.48\textwidth]{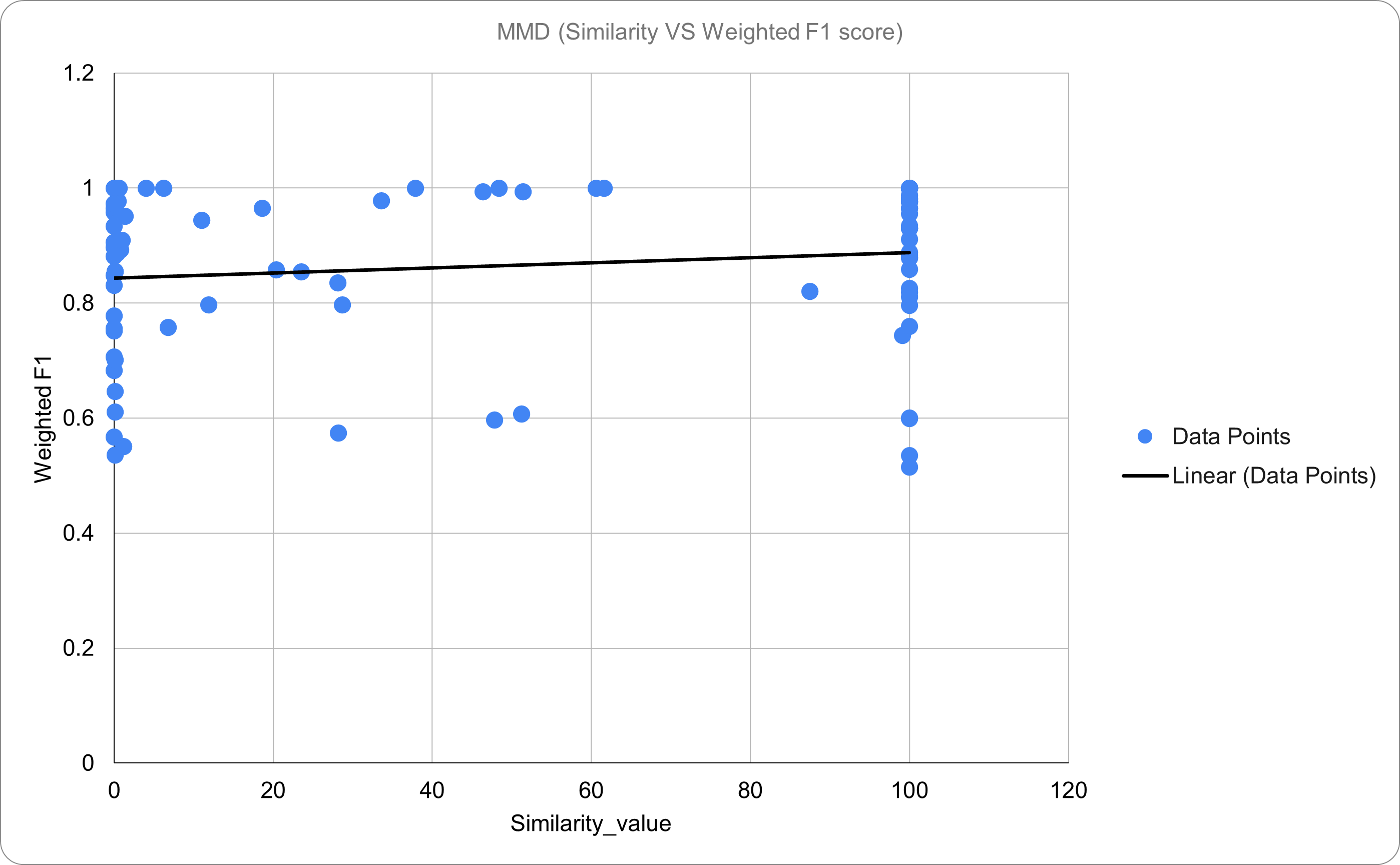}
\caption{Relationship between MMD similarity score and downstream model 
performance across all dataset--method combinations. 
Left: accuracy; Right: weighted F1.}
\label{fig:similarity_performance}
\end{figure}

The results support a moderate positive association between statistical 
similarity and reliable model evaluation. Methods with consistently high 
MMD similarity, including Random, Stratified, and Optimised-Distribution, 
produce broadly convergent performance estimates, while the widest outcome 
variation is concentrated among the lowest-similarity methods, 
Kennard--Stone and SPXY. The bidirectional instability observed for these 
methods, with inflated estimates on easy datasets and deflated estimates on 
harder ones, suggests that highly dissimilar splits reflect structural 
biases of the splitting algorithm rather than the true difficulty of the 
task. As shown in Fig.~\ref{fig:similarity_performance}, the spread of 
outcomes is widest at low similarity values (near zero), where 
Kennard--Stone and SPXY concentrate, and narrows at high similarity values 
(near 100\%), consistent with the positive but weak linear trend visible 
in both plots.

This relationship is moderated by dataset characteristics. On linearly 
separable datasets such as UCI~53, 419, and 857, the choice of splitting 
method has minimal downstream impact regardless of similarity. On datasets 
with low instance-to-feature ratios, such as UCI~74 with 2.8 instances per 
feature, or complex class distributions, such as UCI~264 and UCI~891, the 
choice of method has a more pronounced effect, confirming that similarity 
optimisation delivers the greatest practical benefit where data availability 
is limited and each observation carries greater statistical weight.

The aggregate performance differences between splitting methods are modest, which is unsurprising. Several datasets in the benchmark suite contain tens of thousands to hundreds of thousands of instances, including UCI~891 (253,680 instances), UCI~367 (102,944 instances), and UCI~222 (45,211 instances). As sample size increases, the empirical distribution of a sufficiently large random subsample is expected to approximate the underlying population distribution. Consequently, for large datasets, random, stratified, and geometry-based splitting methods can all produce partitions with similar marginal and joint feature distributions, leaving limited room for similarity-optimised splitting to yield a measurable advantage.

The practical benefit of the Optimised-Distribution method is therefore likely to be greater for smaller datasets or datasets with a low instance-to-feature ratio, such as UCI~42 (214 instances), UCI~857 (200 instances), and UCI~74 (2.8 instances per feature). In these settings, sampling variability is higher, and a poorly chosen split can more substantially distort the estimated data distribution. This also helps explain the ceiling effect observed in Table~\ref{tab:downstream_performance}: on large datasets, the methods converge to similar performance estimates not because the splitting strategy is irrelevant, but because large sample sizes already reduce distributional differences between partitions.

To investigate this further, the 15 datasets were divided into two groups 
based on their average performance under random splitting: a 
\emph{performance-sensitive} group (UCI~42, 45, 74, 222, 367, 519, 891) 
comprising datasets where random splitting yields mean F1 below 0.85, and 
a \emph{performance-stable} group (UCI~17, 53, 75, 159, 264, 419, 529, 857) 
where random splitting already achieves mean F1 of 0.85 or above.
Table~\ref{tab:group_performance_full} summarises average performance, 
standard deviation, and gain relative to random splitting for each group.

\begin{adjustwidth}{-1.5cm}{-1.5cm}
\begin{table}[H]
\centering
\small
\setlength{\tabcolsep}{3pt}
\renewcommand{\arraystretch}{0.95}
\caption{Average downstream performance across dataset groups with standard deviation and gain relative to random splitting.}
\label{tab:group_performance_full}
\begin{tabular}{llrrrrrr}
\toprule
\textbf{Group} & \textbf{Metric} & \textbf{Random} & \textbf{Stratified} & \textbf{K-Stone} & \textbf{SPXY} & \textbf{Duplex} & \textbf{Optimised} \\
\midrule
Performance-sensitive & Accuracy (avg)       & 0.738      & 0.772 & 0.809 & 0.792 & 0.742 & 0.779 \\
          & Accuracy (std)       & 0.118      & 0.131 & 0.153 & 0.159 & 0.114 & 0.120 \\
          & Accuracy avg. gain   & (baseline) & 0.034 & 0.071 & 0.054 & 0.004 & 0.041 \\
          & F1 (avg)             & 0.729      & 0.765 & 0.796 & 0.778 & 0.739 & 0.772 \\
          & F1 (std)             & 0.135      & 0.144 & 0.152 & 0.169 & 0.118 & 0.126 \\
          & F1 avg. gain         & (baseline) & 0.036 & 0.067 & 0.049 & 0.010 & 0.044 \\
\midrule
Performance-stable & Accuracy (avg)     & 0.975      & 0.956 & 0.930 & 0.949 & 0.945 & 0.963 \\
            & Accuracy (std)     & 0.040      & 0.043 & 0.105 & 0.083 & 0.066 & 0.043 \\
            & Accuracy avg. gain & (baseline) & $-$0.018 & $-$0.044 & $-$0.026 & $-$0.030 & $-$0.011 \\
            & F1 (avg)           & 0.974      & 0.956 & 0.925 & 0.947 & 0.945 & 0.963 \\
            & F1 (std)           & 0.041      & 0.044 & 0.107 & 0.087 & 0.066 & 0.043 \\
            & F1 avg. gain       & (baseline) & $-$0.019 & $-$0.049 & $-$0.027 & $-$0.030 & $-$0.011 \\
\bottomrule
\end{tabular}
\end{table}
\end{adjustwidth}

Table~\ref{tab:per_dataset_performance} presents the full per-dataset breakdown, 
illustrating the individual results that underlie the group-level averages.

\begin{table}[H]
\centering
\scriptsize
\setlength{\tabcolsep}{3pt}
\renewcommand{\arraystretch}{0.95}
\caption{Per-dataset downstream performance by splitting method. Datasets are grouped into performance-sensitive (top) and performance-stable (bottom) groups, separated by the horizontal rule.}
\label{tab:per_dataset_performance}
\begin{tabular}{rlrrrrrr}
\toprule
\textbf{UCI} & \textbf{Metric} & \textbf{Random} & \textbf{Stratified} & \textbf{K-Stone} & \textbf{SPXY} & \textbf{Duplex} & \textbf{Optimised} \\
\midrule
42  & Accuracy & 0.754 & 0.831 & 0.815 & 0.879 & 0.758 & 0.831 \\
    & F1       & 0.744 & 0.826 & 0.777 & 0.855 & 0.758 & 0.818 \\
45  & Accuracy & 0.578 & 0.589 & 0.656 & 0.604 & 0.626 & 0.633 \\
    & F1       & 0.514 & 0.535 & 0.646 & 0.550 & 0.607 & 0.600 \\
74  & Accuracy & 0.860 & 0.930 & 0.895 & 0.951 & 0.931 & 0.930 \\
    & F1       & 0.859 & 0.930 & 0.896 & 0.951 & 0.930 & 0.930 \\
222 & Accuracy & 0.859 & 0.863 & 0.940 & 0.906 & 0.752 & 0.859 \\
    & F1       & 0.855 & 0.858 & 0.933 & 0.906 & 0.751 & 0.854 \\
367 & Accuracy & 0.576 & 0.597 & 0.540 & 0.567 & 0.610 & 0.602 \\
    & F1       & 0.574 & 0.597 & 0.535 & 0.567 & 0.610 & 0.599 \\
519 & Accuracy & 0.767 & 0.822 & 0.889 & 0.911 & 0.833 & 0.822 \\
    & F1       & 0.760 & 0.821 & 0.887 & 0.909 & 0.835 & 0.810 \\
891 & Accuracy & 0.772 & 0.773 & 0.927 & 0.723 & 0.685 & 0.773 \\
    & F1       & 0.796 & 0.786 & 0.897 & 0.706 & 0.683 & 0.796 \\
\midrule
17  & Accuracy & 0.977 & 0.936 & 0.960 & 0.965 & 0.977 & 0.983 \\
    & F1       & 0.977 & 0.934 & 0.958 & 0.965 & 0.977 & 0.982 \\
53  & Accuracy & 1.000 & 0.911 & 1.000 & 1.000 & 0.956 & 0.911 \\
    & F1       & 1.000 & 0.911 & 1.000 & 1.000 & 0.956 & 0.911 \\
75  & Accuracy & 0.979 & 0.976 & 0.907 & 0.973 & 0.944 & 0.976 \\
    & F1       & 0.978 & 0.976 & 0.893 & 0.973 & 0.944 & 0.976 \\
159 & Accuracy & 0.881 & 0.881 & 0.873 & 0.769 & 0.831 & 0.884 \\
    & F1       & 0.879 & 0.879 & 0.848 & 0.756 & 0.831 & 0.882 \\
264 & Accuracy & 0.965 & 0.964 & 0.700 & 0.881 & 0.854 & 0.964 \\
    & F1       & 0.965 & 0.964 & 0.701 & 0.881 & 0.854 & 0.964 \\
419 & Accuracy & 1.000 & 1.000 & 1.000 & 1.000 & 1.000 & 1.000 \\
    & F1       & 1.000 & 1.000 & 1.000 & 1.000 & 1.000 & 1.000 \\
529 & Accuracy & 0.994 & 0.981 & 1.000 & 1.000 & 0.994 & 0.987 \\
    & F1       & 0.994 & 0.981 & 1.000 & 1.000 & 0.994 & 0.987 \\
857 & Accuracy & 1.000 & 1.000 & 1.000 & 1.000 & 1.000 & 1.000 \\
    & F1       & 1.000 & 1.000 & 1.000 & 1.000 & 1.000 & 1.000 \\
\bottomrule
\end{tabular}
\end{table}
Tables~\ref{tab:group_performance_full} and~\ref{tab:per_dataset_performance} show that
the effect of splitting strategy is highly dataset-dependent. In the performance-sensitive group, random splitting produces the lowest average 
accuracy (0.738) and weighted F1 (0.729) of all methods. Kennard--Stone achieves 
the strongest average accuracy (0.809, a gain of $+$0.071 over random) and F1 
(0.796, $+$0.067), followed by SPXY with accuracy 0.792 ($+$0.054) and F1 0.778 
($+$0.049). The Optimised-Distribution method achieves a more modest gain of 
$+$0.041 accuracy and $+$0.044 F1, while Duplex provides only marginal improvement 
($+$0.004 accuracy, $+$0.010 F1). This suggests that, for some datasets, broader feature-space coverage can improve model learning even when the resulting train--test partitions are less distributionally similar under MMD. 

In contrast, the performance-stable group exhibits much higher baseline performance 
under random splitting (accuracy 0.975, F1 0.974), leaving little room for 
improvement. All advanced methods \emph{reduce} performance relative to random in 
this group: Kennard--Stone shows the largest drop ($-$0.044 accuracy, $-$0.049 F1), 
followed by Duplex ($-$0.030), SPXY ($-$0.026), Stratified ($-$0.018), and 
Optimised-Distribution ($-$0.011). The notably higher standard deviations for geometry-based methods in this group, such as Kennard--Stone's 0.105 compared with random's 0.040, further indicate that space-filling selection introduces instability when datasets are already easily separable.

This distinction helps explain why the aggregate results in Table~\ref{tab:downstream_performance} are modest despite large differences in statistical similarity. Splitting strategy matters most when the dataset is structurally difficult, moderately sized, or sensitive to feature-space coverage. When the dataset is already easily separable or sufficiently large, random and stratified splits may be adequate because the empirical train and test distributions are already similar enough for stable evaluation. These findings therefore support RQ5: dataset properties moderate the relationship 
between statistical similarity and downstream AutoML performance. Taken together 
with the similarity results in Table~\ref{tab:stat_similarity_scores}, they also 
provide a partial answer to RQ4: improved statistical similarity translates into 
more reliable performance estimates primarily in the performance-sensitive regime, 
where distributional differences between splits have tangible consequences for 
model evaluation. In the performance-stable regime, all methods, regardless of their similarity score, converge to the same ceiling, making the choice of splitting strategy largely inconsequential.

\section{Conclusions and Future Works}
This paper has demonstrated that train--test splitting is a methodological choice with meaningful consequences for the reliability, fairness, and reproducibility of machine learning evaluation. Five established splitting strategies and a proposed Optimised-Distribution refinement were compared systematically across fifteen benchmark datasets using a unified distributional similarity framework. The key findings are:
\begin{itemize}
    \item Splitting methods differ substantially in distributional similarity: mean MMD scores range from 6.9\% for Kennard--Stone to 89.0\% for Optimised-Distribution, a spread of 82.1 percentage points.
    \item Advanced geometry-based methods do not universally outperform random splitting. Kennard--Stone and SPXY consistently produce less similar partitions under the MMD measure due to their space-filling design objectives.
    \item The proposed Optimised-Distribution method achieves the highest distributional similarity of all evaluated strategies, confirming that explicitly treating similarity as an optimisation objective produces measurable improvements.
    \item Improved statistical similarity is associated with more stable and consistent  downstream performance estimates, particularly on structurally complex or low-instance-to-feature-ratio datasets, and most pronouncedly in the performance-sensitive regime identified in Section~\ref{sec:results}.
\end{itemize}

The scope of this work is limited to classification tasks evaluated with a single random seed. Although the evaluation spans 15 datasets with diverse characteristics, including varying sample size, dimensionality, class balance, and domain, within-dataset variance across seeds was not characterised. The findings are nonetheless particularly relevant to small-sample settings, such as clinical and biomedical applications, where dataset sizes are frequently constrained by patient availability or annotation cost, and where a single poorly chosen split can substantially misrepresent true model performance. Future work should therefore replicate the experiments across multiple seeds to confirm the stability of method rankings on individual datasets. 

A parallel open question is characterising which dataset properties determine whether similarity-optimised splitting yields substantial gains. We conjecture that such gains depend not only on sample size and dimensionality, but also on the effective number of informative features. Redundant or recomputable features may artificially inflate dimensionality without contributing meaningfully to the classification boundary, thereby distorting the notion of distributional similarity itself. Identifying these conditions would constitute a self-contained research programme. 

The analysis should further be extended to regression settings where SPXY's continuous target-distance criterion may be more beneficial, and scalable similarity measures, such as random Fourier feature approximations of MMD~\cite{rahimi2007random}, should be explored to reduce the quadratic computational cost. Directly integrating similarity-optimised splitting into AutoML frameworks such as PyCaret and Auto-sklearn would be a natural practical extension.

\bibliographystyle{plain}
\bibliography{biblio}
\end{document}